\documentclass[a4paper,conference]{IEEEtran}
\usepackage{graphicx}
\usepackage{comment}
\usepackage{amssymb,amsmath,amsthm} 
\usepackage{color}

\usepackage{graphicx,multirow,subfigure,comment,algorithm,algorithmic} 
\usepackage[pagebackref=true,breaklinks=true,letterpaper=true,colorlinks,bookmarks=false]{hyperref}
\newcommand{\ci}{~\cite}
\newtheorem{theorem}{Theorem}

\usepackage[table]{xcolor}
\usepackage{caption,booktabs}
\def\BibTeX{{\rm B\kern-.05em{\sc i\kern-.025em b}\kern-.08em
    T\kern-.1667em\lower.7ex\hbox{E}\kern-.125emX}}
\newif\ifimportant
\importanttrue 

\begin{document}
\title{Locality-Promoting Representation Learning}
\author{\IEEEauthorblockN{Johannes Schneider}
\IEEEauthorblockA{Institute of Information Systems, \\
University of Liechtenstein, Vaduz,Liechtenstein \\
johannes.schneider@uni.li}

}

\maketitle 
\ifimportant
\fi

\begin{abstract}
	This work investigates questions related to learning features in convolutional neural networks (CNN). Empirical findings across multiple architectures such as VGG, ResNet, Inception and MobileNet indicate that weights near the center of a filter are larger than weights on the outside. Current regularization schemes violate this principle. Thus, we introduce Locality-promoting Regularization, which yields accuracy gains across multiple architectures and datasets. We also show theoretically that the empirical finding could be explained by maximizing feature cohesion under the assumption of spatial locality. 
\end{abstract}

\ifimportant
\fi

\section{Introduction} 
While the design of deep learning architectures is still a very active field of research, feature engineering seems to be pass\'{e} thanks to deep learning's capability for end-to-end learning. Feature hierarchies are learnt almost miraculously during the optimization of the loss function. Questions such as \emph{``What areas (or patterns) of objects constitute good features?''}, eg. in terms of generalization capability or robustness, have received fairly little attention. In this work, we shed some light on properties and learning of individual features. More precisely, we investigate how weights of features in spatial dimensions of CNNs should be distributed and how to regularize them accordingly. To this end, we focus on spatial data relying on empirical observations and well-known principles, ie. the Principle of Locality. Locality is a known theme in physics and in computer science, eg.\ci{bar16}. In machine learning, CNNs or other fundamental primitives such as word-vectors\ci{mik13} use some form of ``locality'', ie. the idea that interaction strength decreases with distance. It justifies ignoring dependencies among data items, if their distances are above a threshold. Thus, ``windows'' or ``patches'' of inputs can be used rather than the entire input. Prior work by Bengio et al.\ci{ben13} has motivated ideas for representation learning based on principles from physics, but not using the Locality Principle.  Our work addresses a call by Lake et al. \cite{lak17} for better grounding of deep learning by using physical principles. Under the assumption of locality and aiming for cohesive features, it can be observed that weights of features in spatial dimensions close to a center are more relevant, ie. larger. The theoretical finding is supported by empirical evidence through investigating learnt features of multiple architectures. Existing L2-regularization schemes regularize all locations equally, thereby counteracting locality by reducing more central weights too much.  Our first locality promoting L2-regularization scheme (``LOCO-Reg'') for fostering locality results in improvements across multiple architectures such as ResNet, VGG and MobileNet. Furthermore, as a by-product, LOCO-Reg provides a new type of architectural element being a compromise between (spatial) $k\times k$ convolutions and $j\times j$ convolutions with $j\neq k$. For example, dependent on the regularization parameters, a $3 \times 3$ convolution using LOCO-Reg might resemble more of a $1\times 1$ convolution or be closer to a $3 \times 3$ convolution with standard L2-regularization. Our second locality promoting L2-regularization scheme (``STRIP-Reg'') applies this idea to recent work \cite{ding19} to replace 1x3 and 3x1 filters with 3x3 filters.  


\begin{figure}[!htbp]
		\centering	\centerline{
		\includegraphics[width=1.0\linewidth]{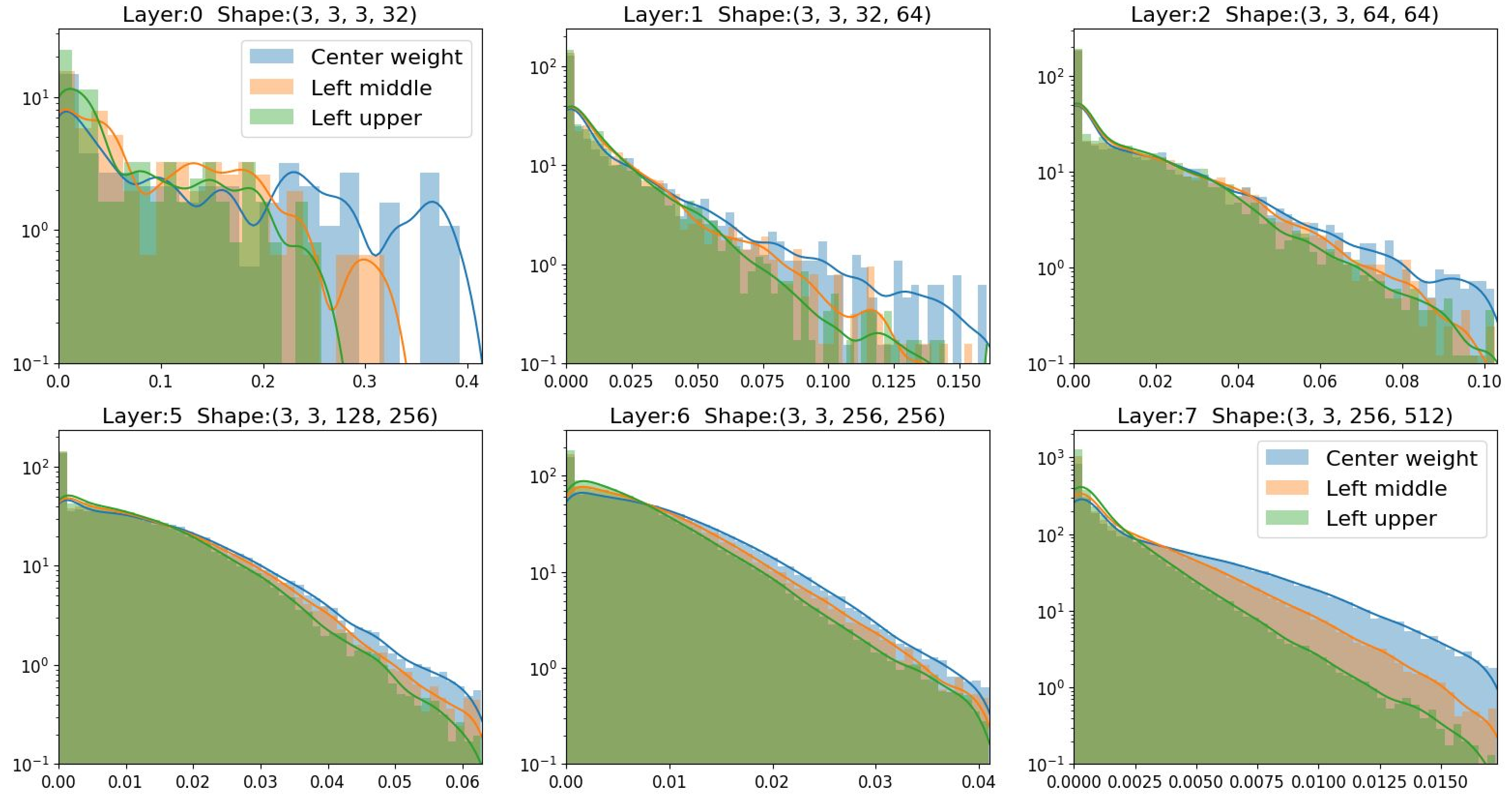}}	
		\caption{Distribution in log-scale of absolute weights of 3x3 filters at center, left middle  and left upper of the first and last 3 conv-layers of a VGG network. Weights near the center tend to be larger.} \label{fig:wdistr}
\end{figure}

\begin{table*}[!htbp]	
	\begin{center}
		\setlength\tabcolsep{2.5pt}
		\begin{tabular}{| l| l  |l|l| l|l| l|l| }\hline
			\multirow{3}{*}{Dataset}& 			\multirow{3}{*}{Architecture}&  \multicolumn{2}{c|}{All Layers}& \multicolumn{2}{c|}{Lower $\frac{1}{2}$ Layers}&\multicolumn{2}{c|}{Upper $\frac{1}{2}$ Layers}\\ \cline{3-8}
			&	&$\overline{I}_{(1,1),(1,0)}$ & $\overline{I}_{(1,0),(0,0)}$& $\overline{I}_{(1,1),(1,0)}$ & $\overline{I}_{(1,0),(0,0)}$&$\overline{I}_{(1,1),(1,0)}$ & $\overline{I}_{(1,0),(0,0)}$\\ \hline
			&VGG16& 0.549\scriptsize{$^{***}$}&0.502\scriptsize{$^{***}$}&0.577\scriptsize{$^{***}$}&0.543\scriptsize{$^{***}$}&0.546\scriptsize{$^{***}$}&0.499\scriptsize{$^{**}$} \\ \cline{2-8} 
			ImageNet&ResNet50  & 0.548\scriptsize{$^{***}$}&0.531\scriptsize{$^{***}$}&0.529\scriptsize{$^{***}$}&0.564\scriptsize{$^{***}$}&0.551\scriptsize{$^{***}$}&0.527\scriptsize{$^{***}$} \\ \cline{2-8}
			\scriptsize{[pre-trained,}&InceptionV3 & 0.49\scriptsize{$^{***}$}&0.447\scriptsize{$^{***}$}&0.573\scriptsize{$^{***}$}&0.532\scriptsize{$^{***}$}&0.483\scriptsize{$^{***}$}&0.439\scriptsize{$^{***}$} \\ \cline{2-8}
			\scriptsize{from Keras]}&Xception & 0.626\scriptsize{$^{***}$}&0.486\scriptsize{$^{***}$}&0.689\scriptsize{$^{***}$}&0.572\scriptsize{$^{***}$}&0.576\scriptsize{$^{***}$}&0.419\scriptsize{$^{***}$} \\ \cline{2-8}
			&MobileNet  & 0.63\scriptsize{$^{***}$}&0.555\scriptsize{$^{***}$}&0.805\scriptsize{$^{***}$}&0.775\scriptsize{$^{***}$}&0.588\scriptsize{$^{***}$}&0.504\scriptsize{$^{}$} \\ \hline
	\end{tabular}\end{center}
	\footnotesize
			\caption{Percentage of spatial filters, where weights in one location are larger than in another. For lower layers, it clearly shows that more central weights are commonly larger. \scriptsize{(*** denotes a p-value $<.001$, ** $<.01$, * $<.1$)}} \label{tab:rat} 
\end{table*}

\section{Spatial Weights Distribution} \label{sec:wei} 
We investigate weights of features along spatial dimensions, that is, width and height of two dimensional images. We denote as a spatial filter, a filter that is of width and height larger one but has just one channel, ie. it is of depth one. Multiple spatial filters constitute a filter, where each spatial filter corresponds to a ``channel''. Weights near a spatial filter's center are on average larger than those near its boundary as illustrated in Figure \ref{fig:wdistr}. We also investigated more formally, if there are significant differences in how often a weight at one position is larger than at another. For a 3x3 spatial filter map $w=(w_{i,j})$ let the indicator $I_{(1,1),(1,0)}$ be 1, if the weight at the center, ie. at location $(1,1)$, is larger than the weight to the left of it, ie. at $(1,0)$. More generally, $I_{(i,j),(k,l)}=1$, if for a $d\times d$ spatial filter $w$ holds $w_{i,j}>w_{k,l}$. We denote by $\overline{I}_{(i,j),(k,l)}$ the average of the indicators for a set of spatial filters $W$. We conduct a binomial-test to investigate whether the average $\overline{I}$ is significantly larger than the expectation, ie. 0.5.
Table \ref{tab:rat} shows the outcome for various architectures.  For lower layers there is consistent support for the hypotheses that weights near the center are larger, but for higher layers it is mostly weaker or there is no support. The overall results is dominated by higher layers, since they contain more features, ie. more spatial filters. One explanation for the discrepancy between lower and higher layers is that down-sampling yields more abstract, semantically richer features with less precise localization. Localization of features in upper layers might be less relevant, since the goal of classification is not localization.  Multiple variations of the test, eg. using different locations or comparing weight magnitudes all yield that more central weights are larger. Overall, we conclude that evidence for locality is likely to be found in any network, but not necessarily on upper layers.  

\section{Locality-Promoting Regularization}  \label{sec:loco}
Current L2-regularization practice might not be optimal. It regularizes weights equally within a spatial filter, thus, battling against the natural tendency for weights in the center being larger than outer ones (Section \ref{sec:wei}). Thus, we aim to encourage learning of weights so that those near the center are larger than those on the outer areas of a spatial filter (A theoretical motivation follows in Section \ref{sec:sep}). One mechanism to guide the learning process is to regularize weights near the center less. To do so, the overall regularization parameter $\lambda$ is adjusted by a factor $r(i,j)$ depending on location $i,j$. For a $d\times d$ spatial filter $w=(w_{i,j})$ the regularization loss becomes $\lambda (\sum_{i,j} r(i,j) ||w_{i,j}||_2)$, for L2-regularization. Note, that for conventional L2-regularization we have $r(i,j)=1$. For simplicity of notation, we focus on $3 \times 3$ spatial filter. Larger filters are considered in the evaluation. We propose two implementations of this idea:

\paragraph{LOCO-Reg} The suggested LOCO-Reg (LOCality-prOmoting Regularization) loss for each spatial filter $w$, corner indexes $I_{co}$ and the four nearest neighbors $I_n$ of the center given by:
\begin{equation} 
\vspace{-3pt}
\footnotesize
\begin{aligned}		
  &\text{LOCO-REG Loss}(w):=
\frac{\lambda}{Z}  \big( w_{1,1}^2+ \gamma\sum_{(i,j) \in I_n}w_{i,j}^2  
  + \eta \sum_{(i,j) \in I_{co}} w_{i,j}^2\big) \label{eq:reg}\\
 &  Z:= \frac{1+4\gamma+4\eta}{9} \text{\phantom{12} Normalization constant}\\
 & I_n:=   \{(0,1),(1,0),(1,2),(2,1)\},   I_{co}:=   \{(0,0),(2,0),(0,2),(2,2)\}  
 \end{aligned}	
\end{equation}


 We use a symmetric function $r$, ie. all corners are regularized identically with parameter $\eta$, all direct neighbors of the center are regularized with parameter $\gamma$. It is convenient to constrain $r(i,j)$, so that $\sum_{i,j} r(i,j)=d^2$. The constraint implies that the meaning of $\lambda$ remains that of the magnitude of the overall regularization. For parameters $\eta=\gamma=1$, we get equal regularization of all weights, ie. standard regularization. LOCO-Reg requires parameters $\eta>\gamma > 1$, which encourages larger weights near the center. To better understand the behavior of LOCO-Reg, or more generally regularization with parameters $r_{i,j}$ that differ spatially, consider the regularization behavior towards the end of the optimization process, ie. once features have been learnt and are not supposed to change significantly. Assume the network learnt an optimal filter $w^*=(w^*_{i,j})$ in terms of generalization performance. The L2-regularization loss is $\lambda \sum_{i,j} r_{i,j} (w^*_{i,j})^2$. One condition that optimal regularization parameters $r^*_{i,j}$ should fulfill is that they should not regularize one of the (optimal) weights of $w^*$ more than another, since this would yield a non-optimal filter. That is, we want for the L2-loss terms that $r^*_{i,j}\cdot (w^*_{i,j})^2=r^*_{k,l}\cdot (w^*_{k,l})^2$ for all $i,j,k,l$. This implies that $r^*_{i,j}\cdot (w^*_{i,j})^2=c_0$ for a constant $c_0$. To determine $c_0$ in $r^*_{i,j} = c_0/ (w^*_{i,j})^2$ we can use the aforementioned constraint that is $\sum_{r,s} r^*(r,s)=d^2$, ie. $c_0:= \frac{d^2}{\sum_{r,s} 1/(w^*_{r,s})^2}$. The formula $r^*_{i,j} = c_0/ (w^*_{i,j})^2$ implies that regularization should be less where weights are larger. Thus, given the empirical observation that learnt (locally) optimal weights of the majority of spatial filter have larger weights (Section \ref{sec:wei}), LOCO-Reg might indeed be preferable to uniform regularization. 

	\begin{figure}[!htbp]
		\small
		\setlength\tabcolsep{2pt}
		\renewcommand{\arraystretch}{1.05}
		\centering
		\begin{tabular}{|c|c|c|c|c|c|c|c|c|c|c|c|c|c|c|c| c|c|} \hline			
			0 & 0 &0 & 0 & 0 &0 & 1 & 1 & 0 & 0 & 1 & 1 & 0 & 0&0&0\\ \hline
			0 & 0 &			1 & 0 & 0 &\cellcolor{red!30}1 & \cellcolor{red!30}2 & \cellcolor{red!30}2 & 1 & 0& \cellcolor{red!15}2 & \cellcolor{red!15}3 & \cellcolor{red!15}1 & 1&1&0\\ \hline
			1 & 0 &			0 & 0 & 1 &\cellcolor{red!30}2 & \cellcolor{violet!30}6 & \cellcolor{violet!30}3 & \cellcolor{blue!30}1& 1 & \cellcolor{red!15}2 & \cellcolor{red!15}5 & \cellcolor{red!15}1 & 2&1&0\\ \hline			
			0 & 0 &			0 & 0 & 1 &\cellcolor{red!30}1 & \cellcolor{violet!30}3 & \cellcolor{violet!30}2 & \cellcolor{blue!30}2 & \cellcolor{blue!15}3& \cellcolor{violet!15}2 & \cellcolor{violet!15}3 & \cellcolor{red!15}2 & 0&0&0\\ \hline
			0 & 1 &			0 & 0 & 0 &1 & \cellcolor{blue!30}2 & \cellcolor{blue!30}2 & \cellcolor{blue!30}2 & \cellcolor{blue!15}2& 3\cellcolor{blue!15} & \cellcolor{blue!15}2 & 2 & 1&0&0\\ \hline			 
			0 & 0 &			0 & 1 & 0 &1 &1 & 1 & 2 & \cellcolor{blue!15}3& \cellcolor{blue!15}3 & \cellcolor{blue!15}2 & 1 & 1&0&0\\ \hline			 			 
		\end{tabular}		
		\caption{Input and localized $3 \times 3$ features. The two red features are more likely discovered by LOCO-Reg and the blue ones more likely with standard regularization.} \label{tab:loc}
\end{figure}

Next, we provide some intuition, on what kinds of spatial filters are fostered using LOCO-Reg. Consider the task of locating two $3 \times 3$ features given just a single feature map shown in Figure \ref{tab:loc}.\footnote{More generally, one might also think of Figure \ref{tab:loc} as a weighted aggregation of feature maps.} Since most values are small, ie. 0 or 1, and potentially superimposed by noise, features seem to be present where the values of the feature maps are large. Figure \ref{tab:loc} shows the outcomes of two strategies for feature definition. The strategy that treats all parts of a proposed feature uniformly places features so that the aggregated sum of inputs covered by the feature is maximal (blue features). The other strategy aligned with LOCO-Reg also seeks areas where the sum is large, but it places more importance on locations near the center (red features). Thus, we might learn features with larger center weights that activate more at maxima of a feature map compared to uniformly regularized features.

\paragraph{STRIP-Reg} We also apply locality-promoting regularization to improve upon a recent architecture that uses the same empirical observation, ie. central weights are larger, to foster locality\cite{ding19}.\footnote{The (first) arxiv.org version of this paper appeared before theirs.} Regularization contributes in an indirect manner to locality in \cite{ding19} as discussed later. We show how locality-promoting L2-regularization might be accommodated in their scheme.  For each 3x3 filter two additional filters are trained in \cite{ding19}: One 1x3 filter applied to the same data as the center row of the 3x3 filter and one 3x1 filter applied to the center column. The outputs of all three filters are added. In architecture ``wBN'' all three filters use separate batchnorm layers before addition, in architecture ``w/oBN'' they use the same batchnorm layer applied after addition.  After training, the weights of all three filters are merged into a single 3x3 filter using the additive property of convolutions. Thus, during inference only a single filter has to be evaluated as for other common architectures. We modify their architecture as follows: We replace the 3x1 and 1x3 filters with 3x3 filters but regularize weights more strongly that are not deemed relevant, ie. that are not present in \cite{ding19}. We call this form of locality promoting regularization, STRIP-Reg, since it regularizes entire horizontal and vertical stripes based on their locality. For the 3x3 filter replacing the 3x1 filter $W_{i,j}$, we regularize a weight $(i,j)$ with $\lambda$ if $(i,j) \in \{(0,1),(0,2),(0,3)\}$ and otherwise by $(1+\eta)\lambda$. The parameter $\lambda$ is the overall regularization strength and $\eta\geq0$ provides the additional regularization of weights not present in \cite{ding19}. 1x3 filters are treated analogously.
We also briefly assess the idea to initialize stronger regularized locations with lower (expected) values, ie. we use a \emph{scaled} initialization: For the 3x3 filter maps replacing the 1x3 and 3x1 maps, we double the weights corresponding to the 1x3 and 3x1 maps and half all other weights after the default initialization, eg. Xavier. This leaves several statistical properties of common initialization schemes in tact, eg. the expectation of the sum of weights of a 3x3 filter remains unchanged for typical initialization schemes like Xavier with independent weight initialization, expectation $E[w_{i,j}]=0$ and standard deviation $E[|w_{i,j}|]=\sigma$. More precisely, $E[\sum_{i,j} a_{i,j}w_{i,j}]=\sum_{i,j} a_{i,j}E[w_{i,j}]=0$, since $E[w_{i,j}]=0$. This also holds for expected magnitudes: $E[\sum_{i,j} a_{i,j}|w_{i,j}|]=\sum_{i,j}a_{i,j}E[|w_{i,j}|]=\sum_{i,j} a_{i,j}\sigma$. We have $a_{i,j}=1$ if all weights are initialized in the same manner, that is $E[\sum_{i,j} a_{i,j}|w_{i,j}|]=9\sigma$. For scaled regularization, 6 weights are initialized with half the expected magnitude $\sigma/2$ and 3 with $2\sigma$. This yields $E[\sum_{i,j} a_{i,j}|w_{i,j}|]=6\sigma/2+ 2\sigma\cdot 3 = 9\sigma$.

While \cite{ding19} provides extensive experimental evaluation supported by a well-structured repo, it lacks any motivation (i) why weights near the center should be larger in general (see Section \ref{sec:sep}), (ii) why their method actually enforces weights near the center to be larger. We discuss point (ii) next.\\
We analyze the network scenario ``w/oBN'', which uses a single batchnorm layer for all three added filters. It is more tractable for analysis: Let $w_{ce}$ be the center weight of a spatial filter and $w_{co}$ a corner weight. \cite{ding19} uses three weights per center of a spatial filter, ie. one for the regular 3x3 convolution, one for the horizontal 1x3 and one for the vertical 3x1 convolution $w_{ce}', w_{ce}'', w_{ce}'''$, while they use just one for a corner $w'_{co}$ originating from the 3x3 convolution. Thus during inference, one might replace $w_{ce}=w_{ce}'+ w_{ce}''+ w_{ce}'''$ and set $w_{co}=w'_{co}$ to obtain the same outcomes as \cite{ding19}.\footnote{See \cite{ding19} for a justification based on linearity of convolutions.}  While the idea to use multiple kernels has several consequences, one is that using three weights instead of one reduces the impact of L2-regularization for central weights. Assume that all weights $w_{ce}', w_{ce}'', w_{ce}'''$ have the same sign and $w_{ce}=w_{ce}'+ w_{ce}'+ w_{ce}'''$ then the regularization loss for a center weight is less if individual weights are regularized before summation, ie. $w_{ce}^2=(w_{ce}'+ w_{ce}''+ w_{ce}''')^2\geq w_{ce}'^2+ w_{ce}''^2+ w_{ce}'''^2$, while the regularization loss is the same for corners $w_{co}^2=w_{co}'^2$. For most initialization the assumption that weights $w_{ce}', w_{ce}'', w_{ce}'''$ have the same sign does not hold. However, let $w_{ce}^*$ be the optimal solution for weight $w_{ce}$ then the optimal the solution for the added weights are $w_{ce}^*=w_{ce}'/3= w_{ce}''/3= w_{ce}'''/3$ since this minimizes the L2-loss, but it does not impact any loss depending on predictions, eg. softmax loss. 

	\begin{figure}[htp]
		
		\includegraphics[width=0.95\linewidth]{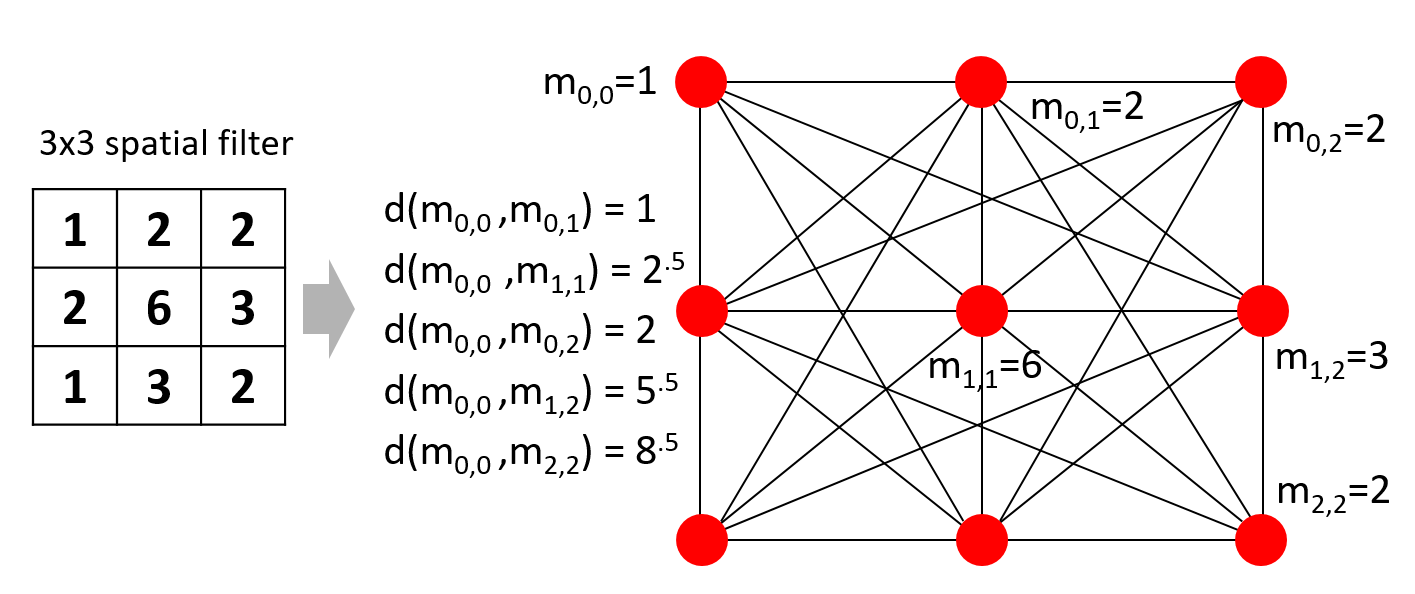}
		\caption{Interpretation of a spatial filter as interacting particles with masses being absolute weights. Particles are red points. Lines indicate interactions. }	\label{fig:locationdist}

\end{figure}
\section{Theory: Feature Cohesion And Locality} \label{sec:sep} 
We show that locality combined with the objective of obtaining cohesive features implies that the more central a location of a spatial filter is, the more relevant it is, ie. the larger the weight should be at that location. As in clustering, ideally features are dense, ie. positioned at areas with high density, and well-separated as captured by cluster assessment metrics like the Davies-Bouldin Index. Cohesion relates naturally to density and it implies that a feature is stable. We seek to define a feature of fixed dimension, ie. $3\times 3$, so that cohesion is maximal.  Our cohesion metric relies on a form of attraction analogous to ``gravity'' that has also been used in the context of clustering \ci{hat11,che02}. We formulate the objective of obtaining highly cohesive features by demanding that feature parts should maximize their attraction. We use the common idealization of point masses, meaning that we neglect the spatial extension of parts of a feature and subsume their strength at a single point (Figure \ref{fig:locationdist}).   Attraction or force between masses at locations $(i,j)$ and $(k,l)$ might be measured analogous to gravity $F(m_{i,j},m_{k,l}):= c_1 \frac{m_{i,j}\cdot m_{k,l}}{d(m_{i,j},m_{k,l})^q}$, where $c_1$ is a constant (as for gravity), $m_{i,j}$ can be seen as the mass or absolute strength of a spatial feature at a location, $d$ is the Euclidean distance and $q>0$ a parameter, eg. we shall use $q=2$ as for gravity. This scenario is illustrated in Figure \ref{fig:locationdist}. 


Cohesion might be measured using the sum of all interactions, i.e. the sum of forces among each pair of parts, ie. $F_{tot}:=\sum_{i,j,k,l} F(m_{i,j},m_{k,l})$ with the force on parts themselves being defined as 0, ie. $F(m_{i,j},m_{i,j}):=0$. We compute distances by taking the differences between indexes, $d(m_{i,j},m_{k,l}):=c_2\sqrt{(i-k)^2+(j-l)^2}$. Any distance is only proportional to the actual (physical) distance with some proportionality constant $c_2$, which can be subsumed in the constant $c_1$ in the force $F(m_{i,j},m_{k,l})$.   

\begin{theorem}
	For any feature strength distribution $m'\leq m_c,m_{co},m_n<(1+\epsilon) m'$ with $\epsilon\in [0,0.675[$, the cohesion $F_{tot}$ of the feature is increased most by increasing $m_c$, and more by increasing any $m_n \in M_n$ than any $m_{co} \in M_{co}$ for arbitrary $m'$, center $m_c=m_{1,1}$, direct neighbors $M_n:= \{m_{1,0},m_{0,1},m_{2,1},m_{1,2}\}$ and corners $M_{co}:= \{m_{0,0},m_{2,0},m_{2,2},m_{0,2}\}$ (Figure \ref{fig:locationdist}).
\end{theorem}
The theorem shows that for many distributions of masses cohesion depends more on masses near the center, ie. increasing any of them yields larger gains in cohesion than masses far away from the center. Locality is incorporated in the definition $F_{tot}$ stating a decrease of interactions with distance. The provided bounds in the theorem might not be tight, but irrespective of this, there are (pathological) cases of feature strength distributions so that enlarging the center might not increase cohesion more than enlarging other parts. For example, assume a corner mass, eg. $m_{0,0}$, is much bigger than any other mass. Then, growing the mass of one of its direct neighbors, eg. $m_{0,1}$, can lead to a more cohesive solution than growing the center $m_{1,1}$. This follows since the center has larger distance to the corner. 


The theorem follows essentially from the fact that the center has (on average) smaller distances to other parts making its contribution to overall cohesion the largest. Analogously, any direct neighbor of the center has (on average) smaller distances to other parts than any corner.

\ifimportant
\begin{proof} 
	We consider the dependency of the total force $F_{tot}$ on the center $m_c$, a corner mass $m_{co}$ and a direct neighbor mass $m_n$ using case enumerations. That is, we investigate the impact on the total force if one of them is changed, while the others stay fixed. Any interaction between masses not involving any of the masses $m_c$, $m_{n}$ or $m_{co}$ can be neglected, since it is not impacted by altering the three masses. Furthermore, due to symmetry only few cases need to be considered.
	
	First, we prove that increasing the center $m_{1,1}$ yields larger change than changing a corner $m_{0,0}$ or direct neighbor $m_{0,1}$. We consider the contribution $F_{c}$ of the center mass $m_{1,1}$ to $F_{tot}$, ie. all interactions that involve $m_{1,1}$ (see Equations \ref{eq:cont}), as well as the contribution $F_{co}$ of the corner mass $m_{0,0}$ and the contribution $F_n$ of the direct neighbor mass $m_{0,1}$. The cases for changing any other mass $m_{i,j}$ are symmetric. The forces can be computed by using the distances shown in Figure \ref{fig:locationdist}. Due to symmetry there are only 4 different distances $1,\sqrt{2},2,\sqrt{5},\sqrt{8}$. 
	
	\begin{equation}
	\footnotesize
	\begin{aligned}		
	F_{c}&:=\sum_{k,l} F(m_{1,1},m_{k,l})  = c_1\cdot m_{1,1}\cdot (\sum_{m \in M_{n}} m +\frac{\sum_{m \in M_{co}} m}{\sqrt{2}^q}) \label{eq:cont} \\
	F_{n}&:= \sum_{k,l} F(m_{0,1},m_{k,l}) =c_1\cdot m_{0,1}\cdot(m_{1,1}+ (\frac{m_{1,0}+m_{1,2}}{\sqrt{2}^q}+\frac{m_{2,1}}{2^q}) 
	\\&+ (m_{0,0}+m_{0,2}+\frac{m_{2,0}+m_{2,2}}{\sqrt{5}^q}))\\
	F_{co}&:= \sum_{k,l} F(m_{0,0},m_{k,l}) =c_1\cdot m_{0,0}\cdot(\frac{m_{1,1}}{\sqrt{2}^q}+\\ &(m_{0,1}+m_{1,0}+\frac{m_{2,1}+m_{1,2}}{\sqrt{5}^q}) + (\frac{m_{2,0}+m_{0,2}}{2^q}+ \frac{m_{2,2}}{\sqrt{8}^q})) \\
	\end{aligned} 
	\end{equation}
	
	Next, we substitute $q=2$ and we compute the impact on the total force when changing one of the three chosen masses, ie. $m_{1,1}$, $m_{0,1}$ and $m_{0,0}$.
	
	\begin{equation}
	\footnotesize
	\begin{aligned}		
	\frac{d F_{c}}{d m_{1,1}} &= c_1\cdot (\sum_{m \in M_{n}} m +\frac{\sum_{m \in M_{co}} m}{2})\\
	\frac{d F_{n}}{d m_{0,1}}&= c_1\cdot (m_{1,1}+ (\frac{m_{1,0}+m_{1,2}}{2}+\frac{m_{2,1}}{4}) + \\&(m_{0,0}+m_{0,2}+\frac{m_{2,0}+m_{2,2}}{5}))\\
	\frac{d F_{co}}{d m_{0,0}}&=c_1\cdot (\frac{m_{1,1}}{2}+ (m_{0,1}+m_{1,0}+\frac{m_{2,1}+m_{1,2}}{5}) +\\
	& (\frac{m_{2,0}+m_{0,2}}{4}+ \frac{m_{2,2}}{8})) \\
	\end{aligned} \label{eq:fo}	
	\end{equation}
	
	We begin with comparing the change of the center to that of the direct neighbor $m_{0,1}$, ie. we start by showing that 
	\begin{equation}
	\footnotesize
	\begin{aligned}		
	\frac{d F_{c}}{d m_{1,1}} > \frac{d F_{n}}{d m_{0,1}}
	\end{aligned} \label{eq:ineq}	
	\end{equation}
	
	We prove inequality (\ref{eq:ineq}) by showing that even under a ``worst-case'' distribution of masses the inequality holds. By assumption for any mass $m_{i,j}$ holds $m_{i,j} \in [m',(1+\epsilon)m']$.  Let us minimize the left hand side ($\frac{d F_{c}}{d m_{1,1}}$) and maximize the right hand side ($\frac{d F_{n}}{d m_{0,1}}$). Formally, we can consider coefficients for masses in Equations \ref{eq:fo}. If a coefficient for a mass in $\frac{d F_{n}}{d m_{0,1}}$ is larger than in $\frac{d F_{c}}{d m_{1,1}}$ then the corresponding mass should be maximized, otherwise minimized. More intuitively, enlarging the mass at location $i,j$ increases $F_{n}$ more than $F_{c}$ if one of the two conditions holds: (i) if location $i,j$ is closer to the neighbor $(0,1)$ than to the center and (ii) if $\frac{d F_{n}}{d m_{0,1}}$ depends on location $i,j$ but $\frac{d F_{c}}{d m_{1,1}}$ does not. Condition (i) only applies to the two corners $(0,0)$ and $(0,2)$ that are nearest to $m_{0,1}$. That is, these two corners should have maximal masses $(1+\epsilon)m'$. For condition (ii) note that the increase of $F_n$ depends on the center $m_{1,1}$ (but not on $m_{0,1}$), whereas the change of $F_{c}$ does not depend on $m_{1,1}$ but on $m_{0,1}$. Thus, the center mass should be maximal, ie. $m_{1,1}=(1+\epsilon) m'$, and all others, including $m_{0,1}$, are minimized, ie. set to $m'$. Substituting the suggested masses into Equations \ref{eq:fo} gives:
	
	\begin{equation*} 
	\footnotesize
	\vspace{-3pt}
	\begin{aligned}	
	&\frac{d F_{c}}{d m_{1,1}} =  c_1\cdot (4m'  + \frac{2m'+ 2(1+\epsilon)m'}{2}) =c_1m'(6+\epsilon)\\
	&\frac{d F_{n}}{d m_{0,1}} =   c_1\cdot((1+\epsilon) m'+ 1.25m' + ((2+\epsilon)m' +0.4m'))\\ 
	& = c_1m'(4.65+3\epsilon)\\
	\end{aligned}	
	\end{equation*} 
	Setting the two terms, ie. $\frac{d F_{c}}{d m_{1,1}}$ and $\frac{d F_{n}}{d m_{0,1}}$, equal gives $\epsilon= 0.675$, ie. the bound is $\epsilon< 0.675$.

	An analogous consideration for the corner mass $m_{0,0}$, ie. investigating if $\frac{d F_{c}}{d m_{1,1}} > \frac{d F_{co}}{d m_{0,0}}$,  yields that in contrast to the direct neighbor $m_{0,1}$, for the corner $m_{0,0}$ there are no masses closer to the corner. Thus, this case is subsumed by the prior case for $m_{0,1}$ stated in Inequality \ref{eq:ineq}, ie. the prior bound $\epsilon< 0.675$ also applies. 
	
	\medskip
	
	Next, we show that changing any of the direct neighbors $M_n$ has greater impact on cohesion, ie. $F_{tot}$, than changing any of the corners $M_{co}$.
	Due to symmetry it suffices to consider only one mass in $M_n$, ie. we chose $m_{0,1}$, and two corners, ie. we use $(0,0)$ and $(2,0)$. We begin by showing that increasing the mass of neighbor $(0,1)$ has more impact than changing the corner $m_{0,0}$. That is, we show $\frac{d F_{n}}{d m_{0,1}}>\frac{d F_{co}}{d m_{0,0}}$. The derivatives are given in Equations \ref{eq:fo}.

	Let us minimize the left hand side ($\frac{d F_{n}}{d m_{0,1}}$) and maximize the right hand side ($\frac{d F_{co}}{d m_{0,0}}$). $\frac{d F_{n}}{d m_{0,1}}$ is minimized and $\frac{d F_{co}}{d m_{0,0}}$ maximized if masses closer to $m_{0,0}$ are maximized or those which only occur with positive coefficient in $\frac{d F_{co}}{d m_{0,0}}$. This means $m_{0,1}$, $m_{1,0}$ and $m_{2,0}$ are chosen to be maximal, ie. they are set to $(1+\epsilon)m'$ and other masses are minimized, ie. set to $m'$.
	
	\begin{equation*} 
	\footnotesize
	\begin{aligned}		
	\frac{d F_{n}}{d m_{0,1}} &=  c_1(m'+0.5m'+0.5(1+\epsilon)m'+0.25m') +  2 m'\\
	&+0.2(1+\epsilon)m'+0.2m')=c_1m'(4.65+0.7\epsilon) \\
	\frac{d F_{co}}{d m_{0,0}} &=  c_1(0.5 m' + 2(1+\epsilon)m'+0.4 m'+0.25(1+\epsilon)m'+ 0.25m'\\
	&+ 0.125m')=c_1m'(3.525+2.25\epsilon) \\
	\end{aligned}	
	\end{equation*} 	
	
	Setting the two terms, ie. $\frac{d F_{n}}{d m_{0,1}}$ and $\frac{d F_{co}}{d m_{0,0}}$, equal gives: $1.125=1.55\epsilon$, ie. $\epsilon=0.725...$.

Finally, we consider the more distant corner $m_{2,0}$. We show $\frac{d F_{n}}{d m_{0,1}}>\frac{d F_{co}}{d m_{2,0}}$. The derivative $\frac{d F_{n}}{d m_{0,1}}$ is given in Equations \ref{eq:fo}. For $\frac{d F_{co}}{d m_{2,0}}$ we have
	\begin{equation*} 
	\footnotesize
	\begin{aligned}		
	\frac{d F_{co}}{d m_{2,0}}&=c_1(\frac{m_{1,1}}{2}+ m_{2,1}+m_{1,0}+\frac{m_{0,1}+m_{1,2}}{5} + \frac{m_{2,2}+m_{0,0}}{4}+ \frac{m_{0,2}}{8})
	\end{aligned}	
	\end{equation*} 
	Let us minimize the left hand side ($\frac{d F_{n}}{d m_{0,1}}$) and maximize the right hand side ($\frac{d F_{co}}{d m_{2,0}}$). As before, $\frac{d F_{n}}{d m_{0,1}}$ is minimized and $\frac{d F_{co}}{d m_{2,0}}$ maximized if masses closer to $m_{0,0}$ are maximized or those which only occur with positive coefficient in $\frac{d F_{co}}{d m_{2,0}}$, ie. $m_{0,1},m_{1,0},m_{2,1}$ and $m_{2,2}$  are maximal, ie. they are set to $(1+\epsilon)m'$ and other masses are minimized, ie. set to $m'$.
	This gives:
	\begin{equation*} 
	\footnotesize
	\vspace{-3pt}
	\begin{aligned}		
	\frac{d F_{n}}{d m_{0,1}} &=  c_1(m'+ (0.5m'+0.75(1+\epsilon)m') +  2 m'+0.2(1+\epsilon)m'+0.2\\
	&=c_1m'(4.65+0.95\epsilon ) \\
	\frac{d F_{co}}{d m_{2,0}} &=  c_1(0.5 m' + 2(1+\epsilon)m'+0.2(1+\epsilon)m'+0.2 m'\\
	&+0.25(1+\epsilon)m'+ 0.25m'+ 0.125m') =c_1m'(3.525+2.45\epsilon) \\
	\end{aligned}	
	\end{equation*} 	
	
	Setting the two terms equal gives $\epsilon= 1.125/1.5=0.75$. Thus, the bound that is valid for all cases is given by $\epsilon< 0.675$.

\end{proof}

\fi

\section{Experiments}
We provide additional empirical support for locality and assess the performance of LOCO-Reg and STRIP-Reg. We investigate different hyper-parameter settings resulting in a total of five experiments (E1-E5) with more than 1000 trained networks.

 \begin{table}[h] 	
 	\vspace{-3pt}
 	\begin{center}
 		\scriptsize
 		\setlength\tabcolsep{2.5pt}
 	
	\centering
 		\begin{tabular}{| l | l| l|l| }\hline
 			\multicolumn{2}{|c|}{MobileNet Adaption\ci{how17}} & \multicolumn{2}{c|}{VGG10 Adaption\ci{sim14}}\\  \hline
			Type/Stride & Filter Shape &Type/Stride& Filter Shape \\ 
 			C/s1     & $3\tiny{\times} 3 \tiny{\times} 3 \tiny{\times} 32$ & C/s1     & $3\tiny{\times} 3 \tiny{\times} 3 \tiny{\times} 32$  \\ \hline
 			C dw/s1     & $3\tiny{\times} 3 \tiny{\times} 32$  & MP/s2     &  $2 \tiny{\times} 2$ \\ \hline
 			C/s1     & $1\tiny{\times} 1 \tiny{\times} 32 \tiny{\times} 32$ & C/s1 & $3\tiny{\times} 3 \tiny{\times} 32 \tiny{\times} 64$  \\ \hline 
 			C dw/s2     & $3\tiny{\times} 3 \tiny{\times} 32$ &  C/s1     & $3\tiny{\times} 3 \tiny{\times} 64 \tiny{\times} 64$ \\ \hline
 			C/s1     & $1\tiny{\times} 1 \tiny{\times} 32 \tiny{\times} 64$ & MP/s2     &  $2 \tiny{\times} 2$ \\ \hline
 			C dw/s1     & $3\tiny{\times} 3 \tiny{\times} 64$ & C/s1 & $3\tiny{\times} 3 \tiny{\times} 64 \tiny{\times} 128$ \\ \hline
 			C/s1     & $1\tiny{\times} 1 \tiny{\times} 64 \tiny{\times} 64$ &  C/s1 &  $3\tiny{\times} 3 \tiny{\times} 128 \tiny{\times} 128$\\ \hline
 			C dw/s2     & $3\tiny{\times} 3 \tiny{\times} 64$ &   MP/s2     & $2 \tiny{\times} 2$\\ \hline
 			C/s1     & $1\tiny{\times} 1 \tiny{\times} 64 \tiny{\times} 128$ &  C/s1 & $3\tiny{\times} 3 \tiny{\times} 128 \tiny{\times} 256$\\ \hline 
 			C dw/s1     & $3\tiny{\times} 3 \tiny{\times} 128$ &  C/s1     & $3\tiny{\times} 3 \tiny{\times} 256 \tiny{\times} 256$ \\ \hline 			
 	
 			C/s1     & $1\tiny{\times} 1 \tiny{\times} 128 \tiny{\times} 128$ &  MP/s2     & $2 \tiny{\times} 2$ \\ \hline
 			C dw/s2     & $3\tiny{\times} 3 \tiny{\times} 128$ &  C/s1 & $3\tiny{\times} 3 \tiny{\times} 256 \tiny{\times} 512$ \\ \hline
 			C/s1     & $1\tiny{\times} 1 \tiny{\times} 128 \tiny{\times} 256$ & C/s1     & $3\tiny{\times} 3 \tiny{\times} 512 \tiny{\times} 512$\\ \hline  
 			C dw/s1     & $3\tiny{\times} 3 \tiny{\times} 256$ &  MP/s2     & $2 \tiny{\times} 2$\\  \hline
 			C/s1     & $1\tiny{\times} 1 \tiny{\times} 256 \tiny{\times} 256$ & &\\  \cline{1-2}
 			C dw/s2     & $3\tiny{\times} 3 \tiny{\times} 256$ & & \\ \cline{1-2}
 			C/s1     & $1\tiny{\times} 1 \tiny{\times} 256 \tiny{\times} 512$ & & \\ \cline{1-2}
 			C dw/s1     & $3\tiny{\times} 3 \tiny{\times} 512$ && \\ \cline{1-2}
 			C/s1     & $1\tiny{\times} 1 \tiny{\times} 512 \tiny{\times} 512$ & & \\ \cline{1-2}
 			C dw/s2     & $3\tiny{\times} 3 \tiny{\times} 512$ && \\ \hline 
 			FC/s1 & $512 \tiny{\times} $nClasses &  FC/s1 & $512 \tiny{\times}$ nClasses \\ \hline
 			Soft/s1 & Classifier &  Soft/s1 & Classifier \\ \hline
 			\end{tabular}
` 	
 	\end{center}
 	\caption{Architectures, where ``C'' is a convolutional,``Soft'' a Softmax and ``MP'' a MaxPool layer; a BatchNorm and a ReLu layer followed each ``C'' layer.}  \label{tab:arch} 
 	\vspace{-6pt}
 \end{table}

\subsection{Setup and Analysis} 
We used SGD with momentum 0.9 and batchsize 128. We trained for 130 epochs decaying the initial learning rate of 0.1 by 0.3 after epochs 50, 85, 110, 120 and 127.  Weights of dense layers were regularized with 0.0005 across all benchmarks. For overall L2-regularization we used $\lambda=0.0005$, if not stated differently.  We used multiple datasets and architectures, where CIFAR-10\ci{kri09} and a VGG-10 variant Table \ref{tab:arch} is the default if not stated differently. We used the datasets default split into training and test data. Data augmentation consisted of horizontal flipping if not stated differently. We trained 15 networks for each configuration, ie. hyperparameter setting.%
We used the Wilcoxon rank-sum test to assess if results differed significantly for two configurations. This test is appropriate for small samples sizes that might contain outliers. We report the median accuracy for each configuration. In all tables bold values indicate best performance comparing values of one column and a set of 2 or 3 rows of differing $\eta,\gamma$. ``***'' denotes a p-value $<.001$, ``**'' denotes a p-value $<.01$, ``*'' a p-value of $<.1$ compared to standard L2-Reg.


\subsection{E1: (Anti-)Locality-Promoting Regularization}
Under the assumption of locality for 3x3 spatial filters regularizing outermost weights more than direct neighbors of the center, ie. $\gamma<\eta$, should yield better results than the opposite, ie. using $\gamma>\eta$. Outcomes for various $\gamma,\eta$ in the upper part of Table \ref{tab:gaet} indeed support this hypotheses:  $\gamma < \eta$ resulted in better accuracies for all tests often with significant differences. Note that in each column of the upper part the center weight is regularized equally, eg. the regularization of the center is the same for $(\gamma,\eta)=(x,y)$ and $(\gamma,\eta)=(y,x)$ for any numbers $x,y$.  We also investigated upon regularizing the center weight more than weights further from it, which can be said to be ``anti''-locality promoting. The results in the lower part of Table \ref{tab:gaet} indicate that regularizing centers more leads to worse outcomes.

	\begin{table}
			\footnotesize 
			\setlength\tabcolsep{3pt}
			\begin{tabular}{|l|l| l|l| l|l| l|l|  }\hline 
			 $\gamma,\eta$	 &Acc. & $\gamma, \eta$	 &Acc.  & $\gamma, \eta$	 &Acc. & $\gamma, \eta$	 &Acc.  \\ \hline 
			1.6,2.4 &	\textbf{.8812}$^{**}$ & 1.6,2.2 &	\textbf{.8793}$^{*}$   & 1.6,2.0 &	\textbf{.8815}$^{*}$       &1.6,1.8 &	\textbf{.8819}   \\ \hline 
			 2.4,1.6&	.874   & 2.2,1.6&	.8781  & 2.0,1.6&	.8789 & 1.8,1.6&	.8789       \\ \hline \hline 
			 
			 1,1&	\textbf{.8790}$^{***}$   & 1,1&	\textbf{.8790}$^{*}$  & 1,1&	\textbf{.8790}$^{**}$ & 1,1&	\textbf{.8790}$^{*}$       \\ \hline
			 .6,.4&	.8677   & .6,.6&	0.8767  & .8,.6&	.8733 & .8,.8&	.8772       \\ \hline
			 
			\end{tabular}
			\caption{Accuracy for $\gamma,\eta$ for $\gamma,\eta$ permutation (upper part) and for (anti)-locality regularization, ie. regularizing distant weights less} \label{tab:gaet}  
		\end{table}

\subsection{E2: LOCO-Reg on Multiple Architectures and Datasets}
We used CIFAR-10, CIFAR-100\ci{kri09} and FASHION-MNIST\ci{xia17}, scaled to 32x32.
We assess three networks types: VGG and MobileNet variants (Table \ref{tab:arch}) as well as ResNet-10\ci{he16}. They cover different design ideas, such as stacking many convolutional layers (VGG), using shortcuts (ResNet) and separating spatial and depth-wise convolutions (MobileNet). 

We chose regularization parameters $\lambda \in \{0.00025,0.0005,0.001,0.002\}$.\footnote{ They were chosen after five runs of standard regularization from a larger set of parameters, so that the best outcome was neither the min or max $\lambda$, ie. neither 0.0025 nor 0.002.} For $(\gamma,\eta)$ we used $\{(1,1), (1.4,1.56), (1.8,2.13)\}$ covering standard regularization $(\gamma,\eta)=(1,1)$ as well as two parameter settings for LOCO-Reg.

\begin{table*}	
	\begin{center}
		\setlength\tabcolsep{2.5pt}
		\footnotesize
		\begin{tabular}{|l|l|l| l|l|l|l|| c|}\hline
\multirow{2}{*}{Dataset}&\multirow{2}{*}{Architecture}&\multirow{2}{*}{($\eta,\gamma$)}&\multicolumn{4}{|c||}{Avg. Accuracy for different $\lambda$}&Best \\ \cline{4-7}
&&& .00025&.0005&.001&.002&Acc. \\ \hline

cifar10&MobileNet&(1,1)&.8611&.8686&.8688&.8647&.8688\\ \hline
cifar10&MobileNet&(1.4,1.56)&.8618&\textbf{.8701$^{*}$}&.8714&.8657&.8714\\ \hline
cifar10&MobileNet&(1.8,2.13)&\textbf{.8619}&.8692&\textbf{.8721$^{*}$}&\textbf{.8668$^{*}$}&\textbf{.8721}$^{*}$\\ \hline\hline
cifar10&ResNet&(1,1)&.9191&.9227&.9236&.9222&.9236\\ \hline
cifar10&ResNet&(1.4,1.56)&\textbf{.921 }&\textbf{.9253$^{*}$}&\textbf{.9242}&.9224&\textbf{.9253}$^{*}$\\ \hline
cifar10&ResNet&(1.8,2.13)&.9186&.9244$^{*}$&.9237&\textbf{.9236}&.9244$^{*}$\\ \hline\hline
cifar10&VGG&(1,1)&.8754&.8761&.882 &.8858&.8858\\ \hline
cifar10&VGG&(1.4,1.56)&.8722&\textbf{.884 $^{**}$}&.8858$^{**}$&.8869&.8869\\ \hline
cifar10&VGG&(1.8,2.13)&\textbf{.8808$^{**}$}&.8816$^{*}$&\textbf{.8875$^{***}$}&\textbf{.8884$^{*}$}&\textbf{.8884}$^{*}$\\ \hline\hline
cifar100&MobileNet&(1,1)&.5926&.6116&.6182&.6155&.6182\\ \hline
cifar100&MobileNet&(1.4,1.56)&\textbf{.5941}&.6124&.6182&.6149&.6182\\ \hline
cifar100&MobileNet&(1.8,2.13)&.5935&\textbf{.6144}&\textbf{.6199}&\textbf{.6184$^{*}$}&\textbf{.6199}\\ \hline\hline
cifar100&ResNet&(1,1)&.702 &.71  &.7156&.7124&.7156\\ \hline
cifar100&ResNet&(1.4,1.56)&.702 &\textbf{.7129$^{*}$}&.7163&\textbf{.7146}&.7163\\ \hline
cifar100&ResNet&(1.8,2.13)&\textbf{.7022}&.7116&\textbf{.7198$^{**}$}&.7142&\textbf{.7198}$^{**}$\\ \hline\hline
cifar100&VGG&(1,1)&.6415&.6551&.6597&.6599&.6599\\ \hline
cifar100&VGG&(1.4,1.56)&.6432&.6583$^{*}$&\textbf{.6665$^{***}$}&.6645$^{*}$&.6665$^{***}$\\ \hline
cifar100&VGG&(1.8,2.13)&\textbf{.6449$^{*}$}&\textbf{.6629$^{***}$}&.6653$^{**}$&\textbf{.6671$^{***}$}&\textbf{.6671}$^{***}$\\ \hline\hline
fashion&MobileNet&(1,1)&\textbf{.9403}&.9402&.939 &.9369&.9403\\ \hline
fashion&MobileNet&(1.4,1.56)&.9398&.9406&.9385&\textbf{.9372}&.9406\\ \hline
fashion&MobileNet&(1.8,2.13)&.9402&\textbf{.9408}&\textbf{.9398}&.9371&\textbf{.9408}\\ \hline\hline
fashion&ResNet&(1,1)&.9501&.9504&.9494&.9492&.9504\\ \hline
fashion&ResNet&(1.4,1.56)&.9496&\textbf{.951 }&.9506$^{*}$&.9489&.951 \\ \hline
fashion&ResNet&(1.8,2.13)&\textbf{.9509$^{*}$}&.9505&\textbf{.9515$^{*}$}&\textbf{.9494}&\textbf{.9515}$^{*}$\\ \hline\hline
fashion&VGG&(1,1)&.9404&\textbf{.942 }&.9417&.9426&.9426\\ \hline
fashion&VGG&(1.4,1.56)&.941 &.9414&.9419&.9436$^{*}$&.9436$^{*}$\\ \hline
fashion&VGG&(1.8,2.13)&\textbf{.9423}&.9417&\textbf{.9436$^{*}$}&\textbf{.9437$^{*}$}&\textbf{.9437}$^{*}$\\ \hline

	\end{tabular}\end{center}
	\caption{Accuracy of LOCO-Reg and standard L2-Reg. LOCO-Reg outperforms almost always, often significantly. \scriptsize{(\textbf{Bold} values indicate best among the 3 values of ($\eta,\gamma$), *** denotes a p-value $<.001$, ** $<.01$, * $<.1$ compared to standard L2-Reg.)}} \label{tab:archRes} 
\end{table*}

Table \ref{tab:archRes} shows results. The last column highlights that LOCO-Reg outperforms on average standard regularization across all datasets and architectures for the best $\lambda$. Gains fluctuate between $0.05\%$ to $0.72\%$. When looking at the outcomes for each $\lambda$ LOCO-Reg outperforms for 34 out of 36 settings (see bold values in Table \ref{tab:archRes}). Often differences are significant despite conducting only 15 runs. 

\subsection{E3: 5x5 and 7x7 spatial filters} \label{sec:E3} We used VGG10 variants based on Table \ref{tab:arch} for CIFAR-100 with convolutions larger than 3x3, ie. we utilized 5x5 as well as 7x7 convolutions. Specifically, we replaced each 3x3 convolution with 5x5 (as well as 7x7) if the convolution did not exceed the feature map's width and height. That is, for VGG$_{5\times5}$ we replaced the first five layers of 3x3 convolutions with 5x5. In VGG$_{7\times7}$ 3x3 filters in the first 3 layers were altered to 7x7 filters and in the 4th and 5th layer we changed to 5x5 filters. We have defined LOCO-Reg using two parameters $\gamma,\eta$ (Equation \ref{eq:reg}). To generalize, we use a simple linear function that gives the regularization parameter given the distance $d$ from the center, ie. 
\begin{equation} 
\vspace{-3pt}
\footnotesize
\begin{aligned}		
u_{a_0,a_1}(d):=a_0+a_1\cdot d \label{eq:rpara}\\
\end{aligned}	
\end{equation}

Outcomes are shown in Table \ref{tab:3x3}. The accuracy gains of LOCO-Reg are profound and highly significant. Since there are more than 2 distances from the center, specifying $\gamma,\eta$ is not sufficient to achieve a gradual regularization from center to the corner. Table \ref{tab:3x3} contains the regularization factors sorted by distance.

	\begin{table}\begin{center}
			\begin{tabular}{| l | l|l | l|l |    }\hline
			Size &	$(a_0,a_1)$ (Eq. \ref{eq:rpara}) &   Acc. & Size &  Acc.  \\ \hline
				5x5           &1,0 &	 .866 &7x7&   .856  \\ \hline
				5x5           &1,0.25   & .873\scriptsize{$^{**}$} &7x7 &   .863\scriptsize{$^{*}$}  \\ \hline 
			5x5           & 1,0.35 &\textbf{.874}\scriptsize{$^{***}$}  &7x7  &  \textbf{.864}\scriptsize{$^{***}$}  \\ \hline 
			\end{tabular}\end{center}
			\caption{Comparison of accuracies for spatial filter sizes 5x5 (VGG$_{5\times5}$) and 7x7 (VGG$_{7\times7}$)} \label{tab:3x3} 
		\end{table}

\subsection{E4: LOCO-Reg on Larger Images}
We also assessed LOCO-Reg on larger images namely 64x64 images (TinyImageNet Dataset\footnote{\url{https://tiny-imagenet.herokuapp.com/}}) and images scaled to 128x128 from the DeepWeeds\cite{Dee19} dataset without data augmentation. We expanded the VGG architecture in Table \ref{tab:arch} by adding a convolutional layer and a max-pooling layer with 32 filters after the first max-pooling layer. Results in Table \ref{tab:lar} indicate that LOCO-Reg yields gains for both datasets.
\begin{table}\begin{center}
\footnotesize
			\begin{tabular}{| l | l|l | l|l |l |    }\hline

				Dataset & ($\gamma,\eta$) &   Acc. & Dataset & ($\gamma,\eta$) &   Acc.  \\ \hline
				
			DeepW. & (1,1) & .73 &TinyIm.& (1,1)  & .4284   \\ \hline
			DeepW. & (1.4,1.56) & .7393\scriptsize{$^{*}$} &TinyIm. &(1.4,1.56) & .4313\scriptsize{$^{*}$}   \\ \hline
			DeepW. & (1.8,2.13) & \textbf{.7484}\scriptsize{$^{*}$} &TinyIm. &(1.8,2.13) & \textbf{.4323}\scriptsize{$^{*}$}   \\ \hline
			\end{tabular}\end{center}
			\caption{Results for 128x128 images (DeepWeeds) and 64x64 images (TinyImageNet) } \label{tab:lar} 
		\end{table}



			

\subsection{E5: STRIP-REG and ACNet}
We first assess the impact of parameter $\eta$. We use $\eta \in \{0,2,8\}$, where $\eta=0$ implies that we use three 3x3 filters that are all treated identically. We used the code provided by \cite{ding19} and trained a Resnet variant ``RCNet-10'' designed for CIFAR-10. We used a predefined learning schedule from \cite{ding19} with SGD training for 80 epochs.

The results in Table \ref{tab:stripReg} indicate that merely using 3x3 filters has no impact. Small amounts of regularization to ``outside'' weights, ie. $\eta=2$,  yield no improvements. Large regularization yield significant improvements compared to the baseline, where scaling of initial values provides a marginal impact. P-values of the Wilcoxon rank-sum test comparing the baseline with STRIP-Reg with $\eta=8$ with default and scaled regularization are 0.05 and 0.04, respectively.

\begin{table}\begin{center}
\footnotesize
			\begin{tabular}{| l | l | l|    }\hline
			Network and Regularization  & Initializa. &  Accuracy  \\ \hline
			ACNet\cite{ding19}(BaseLine) &  Default &  86.18 \\ \hline
			ACNet with 3x3 filters, Default-Reg, $\eta=0$ & Default & 86.20 \\ \hline
			ACNet with 3x3 filters, STRIP-Reg, $\eta=2$  & Default & 86.24\\ \hline
			ACNet with 3x3 filters, STRIP-Reg, $\eta=8$  & Default & 86.36 \\ \hline
			ACNet with 3x3 filters, STRIP-Reg, $\eta=8$ & Scaled &  86.40\\ \hline 
			\end{tabular}\end{center}
			\caption{Results for STRIP-Reg } \label{tab:stripReg} 
		\end{table}

\section{Related Work}
Some priors for ``what makes a good representation'' have been briefly motivated by physics\ci{ben13}. Locality or the idea of emphasizing the center of a representation has not been discussed. The closest weakly related idea is spatial coherence, which is said to imply slow changes of features across spatial dimensions. Locality implies that relevance of feature weights decrease with distance from the center. Disentangling of features is also an important aspect in representation learning. Works on feature disentanglement\ci{tho18,rod16} often constrain representations, eg. weight matrices might be enforced to be orthogonal\ci{rod16}. These works aim at reducing the number of similar representations. Our approach does not directly constrain representations to be dissimilar. Since interactions decrease with distance, larger spatial separation might imply more differences in inputs. Thus, one might hope for obtaining more distinct feature representations compared to the ones that are learnt using patches of nearby or overlapping data. 

Precise location of features has not been deemed essential in the early stages of deep learning. Early works\ci{mut06} on image recognition as well as more recent architectures such as inception\ci{sze15} or VGG\ci{sim14} might use max-pooling that neglects spatial information by simply extracting the maximum out of a region without keeping any information on its position within the considered region. However, it has been recognized that operations such as convolutions\ci{spr14} or fractional max-pooling\ci{gra14} that allow to maintain spatial information more accurately are more suitable for down-sampling. Pooling has been seen as a mechanism to achieve (more) translation invariance. More elaborate approaches such as spatial transformer networks\ci{jad15} allow to learn multiple transformations such as shear, translation and scaling. There is a significant body of work that discusses invariance, e.g.\ci{coh16,bie17}.\ci{coh16} showed how to exploit groups of symmetries such as rotations and reflections. These works aim at replacing the learning of (many) transformable representations, by learning the transformation itself and (fewer) representations on which the transformation can be applied. In contrast, we are more concerned with defining individual representations well rather than reducing the number of representations.

Our work is also loosely related to aspects of the human visual system. Lateral inhibition is an effect, where neurons suppress their less-active neighbors\ci{mut06,pes96}. For instance, it plays a key role for Mach bands, where they increase the contrast between different tones of gray, ie. neurons perform high pass filtering. Our work is neither limited to high pass filtering nor do we strongly inhibit features. But for our implementation we also train features by slightly reducing the impact of outer parts of a feature, ie. the sub-features of which it is composed, to compensate for the promotion of the center.

There are many regularization schemes, such as L1- and L2-regularization, elastic nets\ci{zou05} or versions of dropout such as dropconnect\ci{wan13}. Other works have also used regularization as a means to achieve desired properties of feature representations, eg. regularization in\ci{rod16} pushes pairs of kernels towards a small cosine metric, ie. towards being orthogonal. In contrast, we focus on regularization of elements of a representation rather than pairs of complete representations.

There are also loose connections to pruning of CNNs, eg.\cite{din18,sin19,li16pr}. However, in pruning typically the goal is to fully remove weights, entire filters or replace filters (eg. 3x3 by 1x1) for computational efficiency with little performance loss, while we are primarily concerned with performance improvements.


\section{Conclusions}
Locations near the centre of spatial filters are more important than weights closer to the boundary, meaning that more central weights of the filters are preferable larger on average -- at least for lower layers. The statement is based on empirical findings that are also aligned with theoretical principles and goals, ie. the Principle of Locality and the goal of obtaining cohesive features. The findings can be leveraged using non-uniform, spatial regularization leading to improvements on multiple architectures.

\bibliography{refs}
\bibliographystyle{IEEEtran} 

\end{document}